\pdfoutput=1

\documentclass[11pt]{article}

\usepackage[final]{acl}

\usepackage{times}
\usepackage{latexsym}
\usepackage{listings}
\usepackage{inconsolata}
\usepackage{tabularx}
\usepackage{amsmath}

\usepackage[T1]{fontenc}

\usepackage[utf8]{inputenc}

\usepackage{microtype}
\usepackage{booktabs}

\usepackage{inconsolata}

\usepackage{graphicx}

\lstdefinestyle{aclpython}{
  language=Python,
  basicstyle=\ttfamily\footnotesize,
  keywordstyle=\bfseries,
  commentstyle=\itshape\color{gray},
  stringstyle=\itshape,
  showstringspaces=false,
  numbers=left,
  numberstyle=\tiny,
  stepnumber=1,
  frame=single,
  breaklines=true,
  columns=fullflexible,
  keepspaces=true,
  xleftmargin=2ex
}

\title{\textsc{SafePassage}: High-Fidelity Information Extraction with Black Box LLMs}

\author{Joe Barrow,
  Raj Patel,
  Misha Kharkovski,
  Ben Davies,
  Ryan Schmitt \\
  Pattern Data \\ \texttt{\{first\}.\{last\}@patterndata.ai}}

\begin{document}
\maketitle

\begin{abstract}
Black box large language models (LLMs) make information extraction (IE) easy to configure, but hard to trust.
Unlike traditional information extraction pipelines, the information ``extracted'' is not guaranteed to be grounded in the document.
To prevent this, this paper introduces the notion of a ``safe passage'': context generated by the LLM that is both grounded in the document and consistent with the extracted information.
This is operationalized via a three-step pipeline, \textsc{SafePassage}, which consists of:  (1) an LLM extractor that generates structured entities and their contexts from a document, (2) a string-based global aligner, and (3) a scoring model.
Results show that using these three parts in conjunction reduces hallucinations by up to 85\% on information extraction tasks with minimal risk of flagging non-hallucinations.
High agreement between the \textsc{SafePassage} pipeline and human judgments of extraction quality mean that the pipeline can be dually used to evaluate LLMs.
Surprisingly, results also show that using a transformer encoder fine-tuned on a small number of task-specific examples can outperform an LLM scoring model at flagging unsafe passages.
These annotations can be collected in as little as 1-2 hours.
\end{abstract}

\begin{figure*}[!ht]
    \centering
    \includegraphics[width=\textwidth]{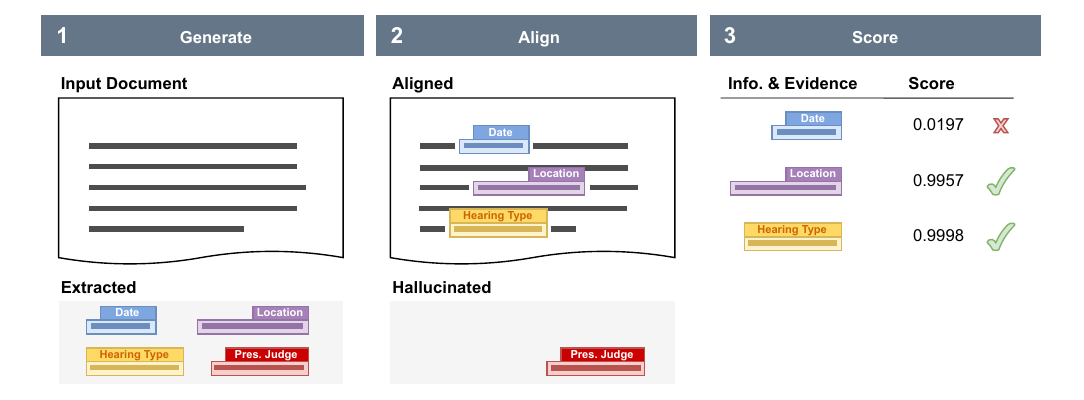}
    
    
    \caption{An overview of the \textsc{SafePassage} pipeline. A generative language model \textit{generates} a list of extracted entities and their contexts, which are \textit{aligned} with the original document, or rejected if they cannot be found in the document. The remaining entities and their aligned contexts are \textit{scored} by some model (e.g. an NLI model or another LLM), and the low-scoring ones are rejected.}
    \label{fig:pipeline}
\end{figure*}

\section{Introduction}
\label{sec:introduction}

Large language models (LLMs) can be used to rapidly and flexibly configure information extraction (IE) pipelines.
With only prompt and schema changes, a black box LLM can be configured across a variety of IE tasks~\citep{brown2020language,liu2023evaluating}.\footnote{We use \emph{black-box} to mean provider-hosted LLMs with no logits, gradients, or attention weights available.} 
Unlike token-level IE systems, however, LLMs provide no \emph{a priori} guarantee that their outputs are grounded in the source document.
The ``extracted'' information could be a hallucination.

This problem is especially dangerous in critical professional settings, such as legal or medical information extraction.
Many law firms are implementing or have implemented generative AI policies, in large part because hallucinations pose professional risks.
Lawyers have been sanctioned for using LLMs to generate briefs that cited fictitious cases~\citep{langham_massachusetts_2024, yang_us_2025}.

With a sufficient harness, LLMs can still be effectively employed as time- and effort-saving tools across the legal domain. 
In order to operationalize such a harness, we propose the concept of a ``safe passage''.
This definition is intentionally model-agnostic: it treats the LLM as a black-box generator of both structured outputs and candidate evidence:

\begin{quote}
\textbf{Definition (safe passage).}
A ``safe passage'' is a snippet copied from the document at generation time that (1) exists in the document, and (2) acts as \textit{supporting evidence} for the extracted information.
\end{quote}

To determine whether a provided context is a ``safe passage'', we propose a \textsc{SafePassage}, a three-step pipeline that requires every extracted field to be accompanied by LLM-copied \textit{evidence} from the document.
The three steps are:
\begin{enumerate}
\itemsep0em
\item \textbf{generate} an entity, a payload, and a context that serves as evidence from the document;
\item \textbf{align} the context with the document, to determine the likelihood it actually occurs in the document; and
\item \textbf{score} the context/evidence against the extracted information to determine if it supports the label.
\end{enumerate}

The resulting score can be used two ways: \emph{prophylactically} as an evaluation for a model or prompt, or \emph{therapeutically} as a set of guardrails to flag hallucinations.
For evaluation, the \textsc{SafePassage} score can be used as a point of comparison between LLM information extraction systems without labeled references.
As guardrails the \textsc{SafePassage} score can be used to rank or suppress individual predictions, or flag them for human review.

We evaluate both LLMs and small encoders as scorers, and find that the small encoders can outperform the LLMs with only a small amount of human annotated data.
The small encoders run at roughly one two-thousandth the cost of the LLM grader, and with substantially lower latency.

Our contributions are:
\begin{enumerate}
\itemsep0em
\item we formalize the concept of a ``safe passage'' for grounded IE with black box LLMs;
\item we introduce \textsc{SafePassage}, a three-stage, model-agnostic pipeline that can serve as both an offline evaluation harness and an online set of guardrails;
\item we demonstrate that a small amount of task-specific labeled data, representing 1-2 hours of annotation effort, can result in a lightweight scorer that outperforms an LLM scorer; and
\item we provide an empirical analysis of \textsc{SafePassage} scores in the legal IE setting, showing that it can flag as many as 85\% of hallucinations.
\end{enumerate}

\section{Related Work}
\label{sec:related_work}








\paragraph{Hallucination Detection for LLMs}
In recent years, LLMs have demonstrated strong performance in a variety of tasks, effectively changing the landscape in many fields.  However, despite their strong performance, they come with certain weaknesses; they sometimes generate text that is false, or at least not backed up by data.  While problematic in most tasks, hallucinations are a particularly large concern in certain critical domains, like legal and medical tasks.  Mistakes and incorrect decisions can lead to severe consequences.  Therefore, detecting and mitigating hallucinations is an important endeavor.  As such, it is a deeply studied area \cite{ji2023survey}.

One major area of study is detecting hallucinations.  \citet{abbes2025small} demonstrates that finetuned encoder-only models are effective at detecting groundedness.  HHEM \cite{hhem-2.1-open} also applies a finetuning approach, this time with Flan-T5.  While shown to be effective, these approaches require finetuning, which may need a large amount of data.  RAGAs \cite{es2024ragas} provides a framework for evaluating RAG \cite{lewis2020retrieval}, which is often used to reduce hallucinations.  RAGAs does not require references; however, this has the downside of focusing on RAG evaluation, not necessarily hallucination evaluation in general.  G-Eval\cite{liu2023g} uses LLMs to detect hallucinations.  Other approaches look at model probabilities or internal LLM states to measure hallucinations \cite{azaria2023internal, yuan2021bartscore}.  These approaches, however, require access to the LLM to detect hallucinations, which may not be possible, depending on the model/framework being used.

\paragraph{LLM Information Extraction and Grounding}
Information extraction is important field in natural language processing.  Much like other fields, transformer based LLMs \cite{vaswani2017attention, devlin2019bert, achiam2023gpt} have been shown to be very effective in IE tasks \cite{brown2020language}.  Like other approaches using LLMs, modern IE approaches are at risk of hallucinations.  In fact, they are a particular concern in information extraction, as the main goal is to extract content present in the input.  Therefore, it is particularly important to emphasis grounding in IE tasks; not just extracting the information, but ensuring it is tied explicitly to somewhere in the input documents.  While LLMs excel at IE tasks, they do not perform well at attributing the solution to its source \cite{liu2023evaluating}. 

In an effort to combat this, approaches have been proposed to attribute LLM IE decisions.  For example, \citet{weller2024according} incorporates prompting strategies to improve attribution.  Approaches like \cite{gao2023enabling, huang2024citation} propose finetuning models explicitly to the attribution task.  Although effective, training costs time and resources, and needs data.  Some approaches like \cite{gao2023rarr, huo2023retrieving},  use the generated content to help determine attribution.  These approaches require additional calls to the LLM, which can be computationally expensive.  Other strategies look at the model during generation to ground output.  \citet{phukan2024peering} focuses on connecting hidden states of important context and generated answers, grounding answers.  This approach does have a downside, however; it requires access to the model, something not guaranteed in the modern LLM landscape.

In this work, we propose \textsc{SafePassage}, an information extraction pipeline that grounds extraction with evidence.  \textsc{SafePassage} has strong results, and the additional advantages of being model agnostic, not requiring extensive training/finetuning, nor requiring access to the internal states of the LLM.  In addition, it does not require additional LLM calls\footnote{An LLM grader can be used (see Section \ref{llm_grader}), but as shown in Section \ref{sec:experiments}, a smaller NLI model is more effective and much cheaper computationally.}.  We discuss \textsc{SafePassage} in the next section.
\section{\textsc{SafePassage} Pipeline}
\label{sec:safepassage}

\begin{figure}[t]
  \centering
  \lstset{style=aclpython}
\begin{lstlisting}
from pydantic import BaseModel
from safepassage import Entity

class Date(BaseModel):
    yyyy: str
    mm: str | None
    dd: str | None

class PredictionDate(Entity):
    date: Date
    context: str

class PresidingJudge(Entity):
    first_name: str
    last_name: str
    context: str

class Response(BaseModel):
    prediction_dates: list[PredictionDate]
    presiding_judge: PresidingJudge
\end{lstlisting}
  \caption{A sample schema for information extraction with LLMs. Each \texttt{Entity} has a \texttt{context} that gets populated with exact text from the document, for later alignment. Using this approach for information extraction allows you to extract structured values, such as the \texttt{Date} composed of year, month, and day, that can be used downstream. Changing which and how many entities are extracted only requires changing the code for entities and the base prompt.}
  \label{fig:response-schema}
\end{figure}

Information extraction with LLMs, particularly with black-box LLMs, introduces new failure modes over traditional approaches that make predictions directly over input tokens. 
In addition to the predictions being \emph{incorrect}, they can be \emph{ungrounded} in the source document.
That is, the extracted information can have no relevant spans or supporting information from the document.

To combat this, we introduce the concept of a 
\textbf{safe passage}, that is, a snippet of evidence  copied by the LLM from the document that (a) exists in the document, and (b) supports the extracted information.
Formally, we define a safe passage as a snippet $c$ (verbatim or near-verbatim) that the LLM copies from the document and that both (a) occurs in the document and (b) \emph{supports} the extracted information $y$ (e.g., an entity or attribute).

If that snippet does not exist in the document, it is considered a hallucination and cannot be used as evidence.
If a snippet exists in the document but does not support the extracted information, it similarly cannot be used as evidence.

To identify safe passages, we introduce a 3-stage guardrails pipeline, shown in Figure~\ref{fig:pipeline}.
\begin{enumerate}
\itemsep0em 
\item \textbf{structured information is generated alongside a \textit{context}} to be used as evidence, where an LLM provides both information and extracted context;
\item \textbf{the context is \textit{aligned} to the contents of the document}, with a string matching algorithm, and discarded if there is no sufficiently good alignment; and
\item \textbf{the aligned context is scored against the generated information}, using either a trained verifier or another LLM, to ensure that the extracted information is supported by the context.
\end{enumerate}

\subsection{Generating Structured Information and Context}

The first step is to have the LLM output candidate passages along with the extracted information, such as entities, entity types, dates, or other structured information.
We make use of structured generation to ensure that the returned values conform to the task-specific schema.
An example of this schema is shown in Figure~\ref{fig:response-schema}.

The \texttt{Entity} schema has three components: a \emph{type}, an optionally \emph{structured payload}  (e.g., \texttt{Date\{yyyy,mm,dd\}}), and (3) a \emph{context}, which is meant to be a span copied from the document.
The structured payloads are a normalized representation that can be used in downstream applications, as opposed to a raw text spans from the document.
The context is reserved for grounding and will be validated in the coming steps to ensure that the extracted entity is in the document.

\subsection{Fuzzy Matching Context to Content}

\begin{figure}[t]
    \centering
    \includegraphics[width=0.5\textwidth]{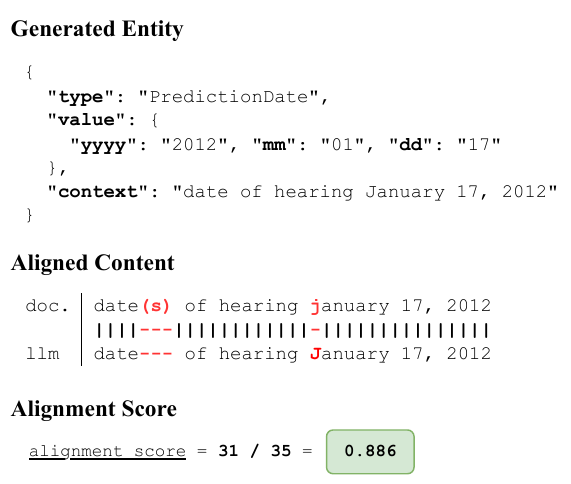}
    
    \caption{Local (Smith–Waterman) alignment of the LLM-provided \emph{context} to document text. Matches are marked with “|”. In this example, the best local alignment has length $L{=}35$ and $M{=}31$ character matches, so the score is $s{=}M/L{=}0.886$.}
    \label{fig:alignment}
\end{figure}

The second stage of the pipeline involves matching each of the contexts to the text content in the document.
If the context is not found in the document, the response is flagged as a potential hallucination and kicked out of the pipeline.

An exact string match does not work with LLM-generated contexts.
Even the best LLMs evaluated in this work had a 5-10\% miss on exact string matches for the context.
The LLMs consistently correct for optical character recognition (OCR) errors, misspellings, and incorrect capitalization.
They also occasionally insert or elide whole words or phrases, or translate words between languages.
Table~\ref{tab:alignment-errors} contains a sample of mistranscriptions encountered in the wild.

We therefore use \emph{local sequence alignment} (Smith–Waterman; \citealp{smith1981identification}) at the character level to retrieve the highest-scoring span $\hat{c}$ from the document for each context $c$. Let the best local alignment have length $L$ and $M$ character matches. We define the alignment score
\begin{equation}
s_{\text{\textsc{align}}}(c,\hat{c}) \;=\; \frac{M}{L} \in [0,1]
\end{equation}

We reject extractions whose context cannot be aligned with sufficient fidelity:
\begin{equation}
\text{keep } (y,c)\ \text{iff}\ s(c,\hat{c}) \ge \tau\quad\text{(we use }\tau{=}0.6\text{).}
\end{equation}
This threshold trades off recall of contexts with some mistranscriptions against precision in excluding hallucinated contexts.
The local alignment in \textsc{SafePassage} is implemented using BioPython~\citep{cock2009biopython}.




\subsection{Evidence Scoring}

After alignment, we have $(y,\hat{c})$: a structured prediction paired with a concrete span from the document. 
The goal is to verify that $\hat{c}$ \emph{supports} $y$.

\subsubsection{Natural Language Inference Scoring}
We cast support checking as a subset of natural language inference (NLI).
In NLI a model predicts a relationship between a pair of texts, a premise, and a hypothesis.
The possible relationships are the premise (a)  \textit{entails}, (b) is \textit{neutral} to, or (c) \textit{contradicts} the hypothesis.

For \textsc{SafePassage}, we format the structured prediction, $y$, into a \textit{hypothesis string}, $h(y)$: \texttt{\{Entity Type\}: \{Entity Value\}} (e.g. "\texttt{Hearing Date: 2012-01-17}").
We use the aligned span from the document as the \textit{premise} (e.g. "\texttt{date(s) of hearing january 17, 2012}").
An NLI model is then used to score the text pairs.

Given $(\hat{c}, h(y))$, we compute a score as the probability of an entailment relationship between the two:
\begin{equation}
s_{\text{NLI}}(y,\hat{c}) \;=\; p(\textsc{entails}\mid \hat{c}, h(y))
\end{equation}

This approach is efficient and cost-effective.
In Section~\ref{sec:results} we detail results using pre-trained NLI models and training models from a small amount of hand-labeled data.

\subsubsection{Using an LLM Grader}
\label{llm_grader}
As an alternative, we propose using an LLM as a grader.
Following~\citet{liu2023g}, we prompt the grader for both a chain of thought and a label.
For evidence scoring, we prompt an LLM grader, $l(y, \hat{c})$, to return a strict label in \{\textsc{supports}, \textsc{insufficient}\} along with some reasoning string.

\begin{equation}
s_{\text{LLM}}(y,\hat{c}, l) \;=\;
\begin{cases}
1 & \text{if } l(h(y), \hat{c}) = \text{\textsc{supports}}, \\
0 & \text{otherwise}.
\end{cases}
\end{equation}

This approach generalizes well to new entity types without retraining, but with higher cost and higher latency.

\section{Experiments}
\label{sec:experiments}

We evaluate \textsc{SafePassage} by measuring the ability of the pipeline to detect hallucinations.
As a basis, we use \textsc{AsyLex} dataset~\citep{barale2023asylex}, which contains Canadian refugee status documents annotated with a variety of entity types.

Unlike the original setting, we consider a mix of structured and unstructured entity types.
\texttt{Date}s and \texttt{DecisionDate}s are structured into separate year, month, and date parts.
\texttt{Hearing Type} is treated as a categorical variable that can take on values \texttt{In Person} or \texttt{Virtual}.
\texttt{Public Or Private Hearing} is also a categorical variable, with values \texttt{Public} or \texttt{Private}
The remaining entity types are unstructured: \texttt{GPE, Organization, Judge}.

\subsection{LLMs Evaluated for Information Extraction}

We benchmark a range of large language models (LLMs) from different providers. From OpenAI, we tested several updated variants of GPT-4~\citep{achiam2023gpt}, including: \texttt{GPT-4.1}, \texttt{GPT-4.1-mini}, \texttt{GPT-4.1-nano}, \texttt{GPT-4o}, as well as a reasoning model, \texttt{o4-mini}.
From Google, we evaluated three variants of Gemini 2.5~\citep{comanici2025gemini}: \texttt{Gemini-2.5-Flash-Lite}, \texttt{Gemini-2.5-Flash} and \texttt{Gemini-2.5-Pro}.
From Meta, we evaluated one Llama 4 model~\citep{grattafiori2024llama}: \texttt{Llama 4 Scout}.
The information extraction prompts and task format are available in the appendices.


\begin{table}[t]
\centering

\begin{tabular}{lccc}
\toprule
\textbf{Scorer} & \textbf{Precision} & \textbf{Recall} & \textbf{F\textsubscript{1}} \\
\midrule
HHEM 2.1 & 0.800 & 0.235 & 0.364 \\
DeBERTAv3 NLI & 0.319 & 0.210 & 0.253 \\
NLI\textsubscript{synth} & 0.937 & 0.705 & 0.804 \\
NLI\textsubscript{human} & \textbf{0.950} & 0.724 & 0.822 \\
NLI\textsubscript{synth>human} & 0.928 & \textbf{0.857} & \textbf{0.891} \\
LLM & 0.878 & 0.782 & 0.827 \\
\bottomrule
\end{tabular}
\caption{Precision (P), Recall (R), and F1 for each scorer. The task is to predict if an extracted entity is a hallucination given the context. The DeBERTaV3 NLI model finetuned on a mix of LLM- and human-annotated data performs best.}
\label{tab:scorers}
\end{table}

\begin{figure*}[!ht]
    \centering
    \includegraphics[width=\textwidth]{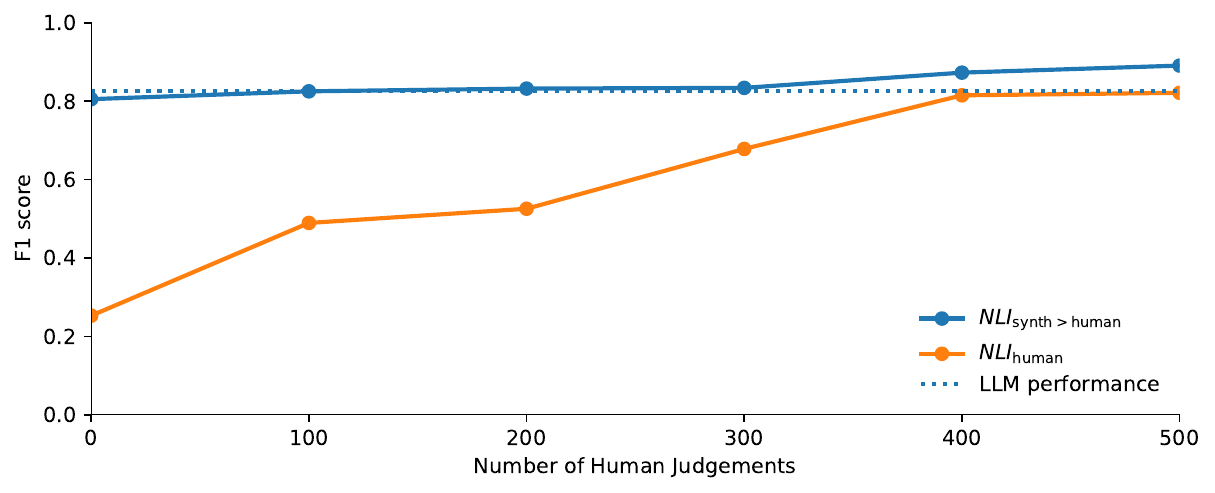}
    
    
    \caption{Finetuned sentence transformer models across dataset sizes. At dataset size 0, the base NLI model is being used without finetuning on human judgments. Pretraining with synthetic data from the LLM scorer improves the NLI scorer at all sizes ($\text{NLI}_{\text{synth>human}}$ vs $\text{NLI}_{\text{human}}$). With just 500 labels (roughly 2 hours of annotation effort) both models perform as well as or better than the LLM scorer.}
    \label{fig:nli_by_counts}
\end{figure*}

\begin{figure*}[!ht]
    \centering
    \includegraphics[width=\textwidth]{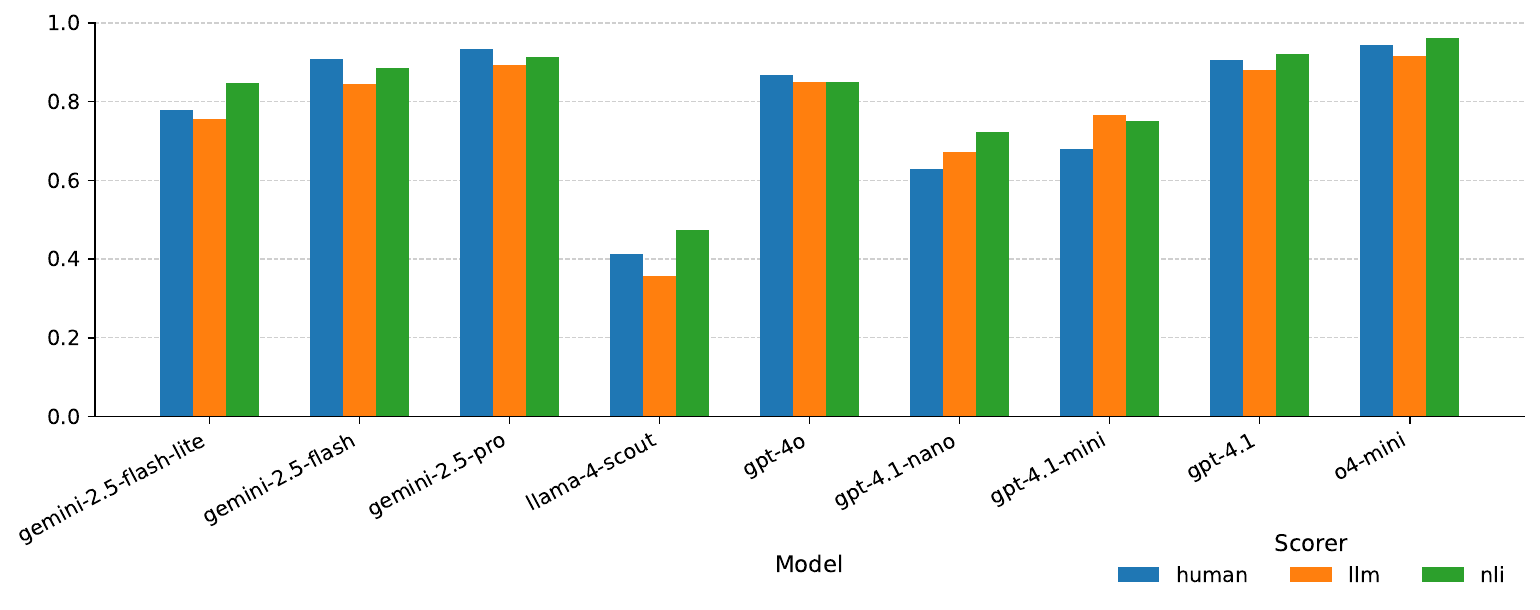}
    
    
    \caption{Average \textsc{SafePassage} scores and average human scores for each of the considered LLMs. Performance trends upward with model size within each group of providers. The best-performing model is the single reasoning model evaluated, \texttt{o4-mini}.}
    \label{fig:model_scores}
\end{figure*}

\subsection{\textsc{SafePassage} Scoring}

We construct a base dataset from the first 1{,}000 documents from \textsc{AsyLex}.
After running an information extraction pipeline across each of the above models, we further annotated 1{,}500 prediction entities across models to construct an entity scoring benchmark.
Each evaluation instance consists of an entity type, the extracted value (structured or unstructured), and the supporting span from the source document.

\subsubsection{Baselines}

We compare two baselines against the \textsc{SafePassage} scorers:
\begin{enumerate}
\itemsep0em
\item \textbf{DeBERTaV3 NLI} DeBERTaV3~\citep{he2021debertav3} is a transformer encoder model, in this case finetuned for NLI as a cross-encoder (both premise and hypothesis fed to the same model).
\item \textbf{HHEM 2.1} The HHEM model~\citep{hhem-2.1-open} is an encoder-decoder model finetuned atop Flan-T5~\citep{chung2024scaling} to detect LLM hallucinations given context and generated text pairs.
\end{enumerate}

For the LLM scorer, we prompt \texttt{GPT-4.1} output a chain-of-thought and a score, as detailed in Section~\ref{sec:safepassage}.
The 3-shot chain-of-thought prompt is available in the appendix.

\subsubsection{Finetuning an NLI Model}
In addition to the baselines, we use up to 500 of human-annotated samples to finetune the DeBERTaV3 NLI model.
This represents \textbf{2 hours of human annotation effort}, measured during annotation.
We also build a silver labeled dataset from 900 unlabeled documents drawn from \textsc{AsyLex}.

\paragraph{Human Annotations}
A human subject matter expert is a source of high-quality gold labels.
Given a simple annotation interface, annotators were able to generate 250 labels an hour consistently.
For training, we used 2 hours of annotation time to generate a set of 500 training samples.

\paragraph{LLM Scorer as a Labeler}
An LLM can be used to efficiently gather a large set of directionally accurate task-specific data. The size of this set depends on the amount of unlabeled data available; for this task we used all the LLM-based \textsc{SafePassage} labels for all models across roughly 900 documents, resulting in roughly 40,000 labels.

\subsubsection{Training and Evaluation}

We use the Sentence Transformers library \citep{reimers-2019-sentence-bert} to finetune the NLI models, using validation $F_1$ as the early stopping criterion.
We used two base models to do the dataset ablations: the DeBERTaV3 NLI model available through the Sentence Transformers library and the same model finetuned on a collection of LLM-generated pseudolabels.
The NLI model finetuned on just the LLM-generated pseudolabels is denoted $\text{NLI}_{\text{synth}}$, the NLI model finetuned on 500 human labels is denoted $\text{NLI}_{\text{human}}$, and the NLI model finetuned on the LLM pseudolabels then subsequently on the 500 human labels is denoted $\text{NLI}_{\text{synth>human}}$.
Results in Table~\ref{tab:scorers} show that the encoder trained on the LLM and task-specific human judgments outperforms all baselines and, surprisingly, the LLM scorer. 

\subsubsection{Gold Dataset Ablation}
In addition to comparing the models, we conduct an ablation study.
We finetuned each base model 5 separate times, once with 100 human labels, 200 human labels, and so on up to 500 human labels.

Experimental results are shown in Figure~\ref{fig:nli_by_counts} and discussed in depth in Section~\ref{sec:results}.
We find that using the LLM-generated silver labels substantially improves the quality of the scorer at all dataset sizes, and allows the small, efficient scorer to outperform the LLM scorer at just 200 human annotations.

\subsection{\textsc{SafePassage} for Comparing LLMs}
This dataset also allows us to compare the information extraction quality of each of the models by comparing their \textsc{SafePassage} scores and their human scores. The results are shown in Figure~\ref{fig:model_scores}, and show that performance trends upward with model size within each group of providers.
The best-performing model is the single reasoning model evaluated, \texttt{o4-mini}.

For this, we compare the human scores from the 1000 predictions, but the \textsc{SafePassage} scores from all predictions made.
We separate out a $\text{\textsc{SafePassage}}_{\text{LLM}}$ and $\text{\textsc{SafePassage}}_{\text{NLI}}$ score, using the best-performing NLI model.

\section{Results \& Analysis}
\label{sec:results}

\begin{table*}[t]
\centering
\small
\setlength{\tabcolsep}{6pt}
\begin{tabularx}{\textwidth}{@{} l X X X l @{}}
\toprule
\textbf{Type} & \textbf{Value} & \textbf{LLM Generated Context} & \textbf{Aligned Context} & \textbf{Error Type} \\
\midrule
Date & \texttt{\{"yyyy": 2013, "mm": 06, "dd": 19\}} & hearing june 19, 2013 & hearing l'audience june 19, 2013 & inserted tokens \\
\midrule
Location & \texttt{InChambers} & (in chambres) & (in chamb ers) & translation/ocr \\
\midrule
Date & \texttt{\{"yyyy": 2007, "mm": 10\}} & October 2007 & o ctober 2007 & ocr \\
\midrule
Date & \texttt{\{"yyyy": 2012, "mm": 07, "dd": 24\}} & date(s) de l'audience july 24, 2012 & date(s) de l'audience february 2, 2012 july 24, 2012 & elided tokens \\
\midrule
Location/GPE & \texttt{Canada} & refugee protection division & refugee protection division & bad grounding \\
\midrule
Public or Private Hearing & \texttt{Private} & date(s) of hearing & date(s) of hearing & bad grounding\\
\midrule
Location/Geopolitical Entity & \texttt{Ontario} & er of citizenship and immigration intimé(e) le ministre de la citoy & er of citizenship and immigration intimé(e) le ministre de la citoy & bad grounding \\
\midrule
Organization & \texttt{iad file no.} & iad file no. & iad file no. & incorrect entity \\
\bottomrule
\end{tabularx}
\caption{Examples of alignment errors between LLM context and aligned context, as well as  hallucinated entities.}
\label{tab:alignment-errors}
\end{table*}

\subsection{\textsc{SafePassage} Improves Hallucination Detection}
The results in Table~\ref{tab:scorers} are the predictive accuracy of each scorer for detecting \emph{unsafe passages}.
When detecting unsafe passages, high recall is important to reduce the likelihood of a hallucination slipping through, and high precision is important to ensure that safe passages are not dropped or filtered.
The $\text{NLI}_{\text{synth>human}}$ scorer achieves a recall of 0.857, which means that ~85\% of unsafe passages (hallucinations) were correctly detected.

\subsection{Two Hours of Labeler Time Can Produce a High-Quality Scorer}
One of the most surprising results of our experiments, building on a result from~\citet{abbes2025small}, is that a small encoder can not only \emph{rival} a large decoder for scoring, but \emph{surpass it}.
Even with a small number of human annotations, in the low hundreds.
In our setting, we specifically consider task-specific labels for training the NLI models.
That is, all entity and context pairs that are labeled are drawn from the same closed set of entities.
At roughly 250 annotations per hour, a training set of 500 annotations takes approximately two hours of human annotation time.

Table~\ref{tab:scorers} and Figure~\ref{fig:nli_by_counts} both support the use of LLM silver labels.
These are labels generated over a moderate number of input documents (900) used to finetune the NLI model \emph{before} finetuning on the human annotations.
Figure~\ref{fig:nli_by_counts} shows that the silver labels improve the accuracy of the NLI model at all considered dataset sizes.

\subsection{Transformer Encoders Are Efficient, High-Quality Scorers}
The biggest benefit of using a small encoder as opposed to an LLM grader is the decrease in latency and cost of scoring with \textsc{SafePassage}.
On a standard T4 GPU, the NLI scorer achieves 250 predictions per second, or less than \$7e-7 per prediction at a standard cloud GPU rental rate (\$0.59/hour).
In contrast, the LLM grader, with an average of 250 input tokens and 100 output tokens, costs roughly \$0.002 per prediction.
This means that the LLM scorer costs roughly 2{,}000 times the NLI scorer per prediction.
Beyond the cost savings, the small encoder scorer runs at substantially lower latency than the LLM grader.

\subsection{Larger LLMs Are Better Information Extractors}
Both the human results and the \textsc{SafePassage} results in Figure~\ref{fig:model_scores} find that the presumably larger models outperform the smaller models within model groups.
For instance, \texttt{GPT-4.1} > \texttt{GPT-4.1-Mini} > \texttt{GPT-4.1-Nano}, and \texttt{Gemini 2.5 Pro} > \texttt{Gemini 2.5 Flash} > \texttt{Gemini 2.5 Flash-Lite}.

Two surprising findings are the strength of the one reasoning model evaluated, \texttt{o4-mini}, and the weakness of \texttt{Llama 4 Scout}.
The entities extracted with Llama showed poor instruction following, which could be partially due to the structured output format.
Previous work has shown that structured outputs can degrade response quality~\citep{tam-etal-2024-speak}.
Exploring how to improve information extraction results with black box LLMs is a worthwhile future line of work.

\subsection{LLMs Do Not Perfectly Copy Their Input}
In addition to the above experiments, we also do qualitative analysis on the NLI model. 
A selection of errors are shown in Table~\ref{tab:alignment-errors}.

The first four errors in the table are mistranscriptions, where the LLM made some change to the surface form of the copied context.
These changes include: remove extraneous spaces, capitalizing words, inserting or eliding words, and in one case translating (chambers/chambres).

Even the best performing LLM in our experiments, \texttt{o4-mini}, had roughly a 5\% rate of mistranscriptions.
The worst performing LLM in our experiment, \texttt{Llama 4 Scout} had some kind of mistranscription for 42\% of the extracted entities.

\subsection{LLMs Can Reliably Copy Useful Evidence}
Although they do not perfectly copy evidence, LLMs still reliably copy \emph{useful} evidence.
Because it makes local predictions, \textsc{SafePassage} relies on the ability of LLMs to output supportive and meaningful evidence.
Figure~\ref{fig:model_scores} shows that ~95\% of the entities that \texttt{o4-mini} predicts are substantiated by grounded evidence in the documents.
On \textsc{AsyLex} \texttt{o4-mini} predicted an average of 9 entities per document, meaning it produced an unsafe passage in only once every 2 documents.
\section{Conclusions}
\label{sec:conclusion}

\textsc{SafePassage} enables the use of black box LLMs for high-fidelity information extraction (IE) by requiring that each extracted field be justified by a \emph{safe passage}: evidence that is both (a) contained in the document (via alignment), and (b) supports the claimed value (via a scorer).
The resulting score can be used offline to evaluate the reliability of IE systems and online to flag potential hallucinations.
In our legal IE setting, we further show that with only a small amount of subject matter expert-annotated data, it is possible to train an efficient transformer encoder as a scorer that outperforms an LLM grader.

\paragraph{Limitations and Future Work}
By design \textsc{SafePassage} improves precision of an IE pipeline by filtering out potential hallucinations.
In the future, we see several possible solutions that can improve \textit{both} recall and precision, such as ensembling calls or using \textsc{SafePassage} scores as a feedback mechanism in a multi-turn setting.
In addition, we showed the effectiveness of a task-specific scorer that is efficient to train and inference.
In future work we aim to improve more task-agnostic scorers.

\bibliography{custom}

\begin{thebibliography}{30}
\providecommand{\natexlab}[1]{#1}

\bibitem[{Abbes et~al.(2025)Abbes, Prato, Fournier, Rodriguez, Boukhary, Elwood, and Chandar}]{abbes2025small}
Istabrak Abbes, Gabriele Prato, Quentin Fournier, Fernando Rodriguez, Alaa Boukhary, Adam Elwood, and Sarath Chandar. 2025.
\newblock Small encoders can rival large decoders in detecting groundedness.
\newblock \emph{arXiv preprint arXiv:2506.21288}.

\bibitem[{Achiam et~al.(2023)Achiam, Adler, Agarwal, Ahmad, Akkaya, Aleman, Almeida, Altenschmidt, Altman, Anadkat et~al.}]{achiam2023gpt}
Josh Achiam, Steven Adler, Sandhini Agarwal, Lama Ahmad, Ilge Akkaya, Florencia~Leoni Aleman, Diogo Almeida, Janko Altenschmidt, Sam Altman, Shyamal Anadkat, and 1 others. 2023.
\newblock Gpt-4 technical report.
\newblock \emph{arXiv preprint arXiv:2303.08774}.

\bibitem[{Azaria and Mitchell(2023)}]{azaria2023internal}
Amos Azaria and Tom Mitchell. 2023.
\newblock The internal state of an llm knows when it’s lying.
\newblock In \emph{Findings of the Association for Computational Linguistics: EMNLP 2023}, pages 967--976.

\bibitem[{Bao et~al.(2024)Bao, Li, Luo, and Mendelevitch}]{hhem-2.1-open}
Forrest Bao, Miaoran Li, Rogger Luo, and Ofer Mendelevitch. 2024.
\newblock \href {https://doi.org/10.57967/hf/3240} {{HHEM-2.1-Open}}.

\bibitem[{Barale et~al.(2023)Barale, Klaisoongnoen, Minervini, Rovatsos, and Bhuta}]{barale2023asylex}
Claire Barale, Mark Klaisoongnoen, Pasquale Minervini, Michael Rovatsos, and Nehal Bhuta. 2023.
\newblock Asylex: A dataset for legal language processing of refugee claims.
\newblock In \emph{The 5th Natural Legal Language Processing Workshop 2023}, pages 244--257. Association for Computational Linguistics (ACL).

\bibitem[{Brown et~al.(2020)Brown, Mann, Ryder, Subbiah, Kaplan, Dhariwal, Neelakantan, Shyam, Sastry, Askell et~al.}]{brown2020language}
Tom Brown, Benjamin Mann, Nick Ryder, Melanie Subbiah, Jared~D Kaplan, Prafulla Dhariwal, Arvind Neelakantan, Pranav Shyam, Girish Sastry, Amanda Askell, and 1 others. 2020.
\newblock Language models are few-shot learners.
\newblock \emph{Advances in neural information processing systems}, 33:1877--1901.

\bibitem[{Chung et~al.(2024)Chung, Hou, Longpre, Zoph, Tay, Fedus, Li, Wang, Dehghani, Brahma et~al.}]{chung2024scaling}
Hyung~Won Chung, Le~Hou, Shayne Longpre, Barret Zoph, Yi~Tay, William Fedus, Yunxuan Li, Xuezhi Wang, Mostafa Dehghani, Siddhartha Brahma, and 1 others. 2024.
\newblock Scaling instruction-finetuned language models.
\newblock \emph{Journal of Machine Learning Research}, 25(70):1--53.

\bibitem[{Cock et~al.(2009)Cock, Antao, Chang, Chapman, Cox, Dalke, Friedberg, Hamelryck, Kauff, Wilczynski et~al.}]{cock2009biopython}
Peter~JA Cock, Tiago Antao, Jeffrey~T Chang, Brad~A Chapman, Cymon~J Cox, Andrew Dalke, Iddo Friedberg, Thomas Hamelryck, Frank Kauff, Bartek Wilczynski, and 1 others. 2009.
\newblock Biopython: freely available python tools for computational molecular biology and bioinformatics.
\newblock \emph{Bioinformatics}, 25(11):1422.

\bibitem[{Comanici et~al.(2025)Comanici, Bieber, Schaekermann, Pasupat, Sachdeva, Dhillon, Blistein, Ram, Zhang, Rosen et~al.}]{comanici2025gemini}
Gheorghe Comanici, Eric Bieber, Mike Schaekermann, Ice Pasupat, Noveen Sachdeva, Inderjit Dhillon, Marcel Blistein, Ori Ram, Dan Zhang, Evan Rosen, and 1 others. 2025.
\newblock Gemini 2.5: Pushing the frontier with advanced reasoning, multimodality, long context, and next generation agentic capabilities.
\newblock \emph{arXiv preprint arXiv:2507.06261}.

\bibitem[{Devlin et~al.(2019)Devlin, Chang, Lee, and Toutanova}]{devlin2019bert}
Jacob Devlin, Ming-Wei Chang, Kenton Lee, and Kristina Toutanova. 2019.
\newblock Bert: Pre-training of deep bidirectional transformers for language understanding.
\newblock In \emph{Proceedings of the 2019 conference of the North American chapter of the association for computational linguistics: human language technologies, volume 1 (long and short papers)}, pages 4171--4186.

\bibitem[{Es et~al.(2024)Es, James, Anke, and Schockaert}]{es2024ragas}
Shahul Es, Jithin James, Luis~Espinosa Anke, and Steven Schockaert. 2024.
\newblock Ragas: Automated evaluation of retrieval augmented generation.
\newblock In \emph{Proceedings of the 18th Conference of the European Chapter of the Association for Computational Linguistics: System Demonstrations}, pages 150--158.

\bibitem[{Gao et~al.(2023{\natexlab{a}})Gao, Dai, Pasupat, Chen, Chaganty, Fan, Zhao, Lao, Lee, Juan et~al.}]{gao2023rarr}
Luyu Gao, Zhuyun Dai, Panupong Pasupat, Anthony Chen, Arun~Tejasvi Chaganty, Yicheng Fan, Vincent Zhao, Ni~Lao, Hongrae Lee, Da-Cheng Juan, and 1 others. 2023{\natexlab{a}}.
\newblock Rarr: Researching and revising what language models say, using language models.
\newblock In \emph{Proceedings of the 61st Annual Meeting of the Association for Computational Linguistics (Volume 1: Long Papers)}, pages 16477--16508.

\bibitem[{Gao et~al.(2023{\natexlab{b}})Gao, Yen, Yu, and Chen}]{gao2023enabling}
Tianyu Gao, Howard Yen, Jiatong Yu, and Danqi Chen. 2023{\natexlab{b}}.
\newblock Enabling large language models to generate text with citations.
\newblock In \emph{The 2023 Conference on Empirical Methods in Natural Language Processing}.

\bibitem[{Grattafiori et~al.(2024)Grattafiori, Dubey, Jauhri, Pandey, Kadian, Al-Dahle, Letman, Mathur, Schelten, Vaughan et~al.}]{grattafiori2024llama}
Aaron Grattafiori, Abhimanyu Dubey, Abhinav Jauhri, Abhinav Pandey, Abhishek Kadian, Ahmad Al-Dahle, Aiesha Letman, Akhil Mathur, Alan Schelten, Alex Vaughan, and 1 others. 2024.
\newblock The llama 3 herd of models.
\newblock \emph{arXiv preprint arXiv:2407.21783}.

\bibitem[{He et~al.(2021)He, Gao, and Chen}]{he2021debertav3}
Pengcheng He, Jianfeng Gao, and Weizhu Chen. 2021.
\newblock Debertav3: Improving deberta using electra-style pre-training with gradient-disentangled embedding sharing.
\newblock \emph{arXiv preprint arXiv:2111.09543}.

\bibitem[{Huang and Chang(2024)}]{huang2024citation}
Jie Huang and Kevin Chen~Chuan Chang. 2024.
\newblock Citation: A key to building responsible and accountable large language models.
\newblock In \emph{2024 Findings of the Association for Computational Linguistics: NAACL 2024}, pages 464--473. Association for Computational Linguistics (ACL).

\bibitem[{Huo et~al.(2023)Huo, Arabzadeh, and Clarke}]{huo2023retrieving}
Siqing Huo, Negar Arabzadeh, and Charles Clarke. 2023.
\newblock Retrieving supporting evidence for generative question answering.
\newblock In \emph{Proceedings of the annual international acm sigir conference on research and development in information retrieval in the Asia Pacific region}, pages 11--20.

\bibitem[{Ji et~al.(2023)Ji, Lee, Frieske, Yu, Su, Xu, Ishii, Bang, Madotto, and Fung}]{ji2023survey}
Ziwei Ji, Nayeon Lee, Rita Frieske, Tiezheng Yu, Dan Su, Yan Xu, Etsuko Ishii, Ye~Jin Bang, Andrea Madotto, and Pascale Fung. 2023.
\newblock Survey of hallucination in natural language generation.
\newblock \emph{ACM computing surveys}, 55(12):1--38.

\bibitem[{Langham(2024)}]{langham_massachusetts_2024}
Pamela Langham. 2024.
\newblock Massachusetts lawyer sanctioned for ai-generated fictitious case citations.
\newblock \url{https://www.msba.org/site/site/content/News-and-Publications/News/General-News/Massachusetts_Lawyer-Sanctioned_for_AI_Generated-Fictitious_Cases.aspx}.

\bibitem[{Lewis et~al.(2020)Lewis, Perez, Piktus, Petroni, Karpukhin, Goyal, K{\"u}ttler, Lewis, Yih, Rockt{\"a}schel et~al.}]{lewis2020retrieval}
Patrick Lewis, Ethan Perez, Aleksandra Piktus, Fabio Petroni, Vladimir Karpukhin, Naman Goyal, Heinrich K{\"u}ttler, Mike Lewis, Wen-tau Yih, Tim Rockt{\"a}schel, and 1 others. 2020.
\newblock Retrieval-augmented generation for knowledge-intensive nlp tasks.
\newblock \emph{Advances in neural information processing systems}, 33:9459--9474.

\bibitem[{Liu et~al.(2023{\natexlab{a}})Liu, Zhang, and Liang}]{liu2023evaluating}
Nelson~F Liu, Tianyi Zhang, and Percy Liang. 2023{\natexlab{a}}.
\newblock Evaluating verifiability in generative search engines.
\newblock In \emph{The 2023 Conference on Empirical Methods in Natural Language Processing}.

\bibitem[{Liu et~al.(2023{\natexlab{b}})Liu, Iter, Xu, Wang, Xu, and Zhu}]{liu2023g}
Yang Liu, Dan Iter, Yichong Xu, Shuohang Wang, Ruochen Xu, and Chenguang Zhu. 2023{\natexlab{b}}.
\newblock G-eval: Nlg evaluation using gpt-4 with better human alignment.
\newblock In \emph{Proceedings of the 2023 Conference on Empirical Methods in Natural Language Processing}, pages 2511--2522.

\bibitem[{Phukan et~al.(2024)Phukan, Somasundaram, Saxena, Goswami, and Srinivasan}]{phukan2024peering}
Anirudh Phukan, Shwetha Somasundaram, Apoorv Saxena, Koustava Goswami, and Balaji~Vasan Srinivasan. 2024.
\newblock Peering into the mind of language models: An approach for attribution in contextual question answering.
\newblock In \emph{Findings of the Association for Computational Linguistics ACL 2024}, pages 11481--11495.

\bibitem[{Reimers and Gurevych(2019)}]{reimers-2019-sentence-bert}
Nils Reimers and Iryna Gurevych. 2019.
\newblock \href {https://arxiv.org/abs/1908.10084} {Sentence-bert: Sentence embeddings using siamese bert-networks}.
\newblock In \emph{Proceedings of the 2019 Conference on Empirical Methods in Natural Language Processing}. Association for Computational Linguistics.

\bibitem[{Smith et~al.(1981)Smith, Waterman et~al.}]{smith1981identification}
Temple~F Smith, Michael~S Waterman, and 1 others. 1981.
\newblock Identification of common molecular subsequences.
\newblock \emph{Journal of Molecular Biology}, 147(1):195--197.

\bibitem[{Tam et~al.(2024)Tam, Wu, Tsai, Lin, Lee, and Chen}]{tam-etal-2024-speak}
Zhi~Rui Tam, Cheng-Kuang Wu, Yi-Lin Tsai, Chieh-Yen Lin, Hung-yi Lee, and Yun-Nung Chen. 2024.
\newblock \href {https://doi.org/10.18653/v1/2024.emnlp-industry.91} {Let me speak freely? a study on the impact of format restrictions on large language model performance.}
\newblock In \emph{Proceedings of the 2024 Conference on Empirical Methods in Natural Language Processing: Industry Track}, pages 1218--1236, Miami, Florida, US. Association for Computational Linguistics.

\bibitem[{Vaswani et~al.(2017)Vaswani, Shazeer, Parmar, Uszkoreit, Jones, Gomez, Kaiser, and Polosukhin}]{vaswani2017attention}
Ashish Vaswani, Noam Shazeer, Niki Parmar, Jakob Uszkoreit, Llion Jones, Aidan~N Gomez, {\L}ukasz Kaiser, and Illia Polosukhin. 2017.
\newblock Attention is all you need.
\newblock \emph{Advances in neural information processing systems}, 30.

\bibitem[{Weller et~al.(2024)Weller, Marone, Weir, Lawrie, Khashabi, and Van~Durme}]{weller2024according}
Orion Weller, Marc Marone, Nathaniel Weir, Dawn Lawrie, Daniel Khashabi, and Benjamin Van~Durme. 2024.
\newblock “according to...”: Prompting language models improves quoting from pre-training data.
\newblock In \emph{Proceedings of the 18th Conference of the European Chapter of the Association for Computational Linguistics (Volume 1: Long Papers)}, pages 2288--2301.

\bibitem[{Yang(2025)}]{yang_us_2025}
Maya Yang. 2025.
\newblock \href {https://www.theguardian.com/us-news/2025/may/31/utah-lawyer-chatgpt-ai-court-brief} {Us lawyer sanctioned after being caught using {ChatGPT} for court brief}.
\newblock \emph{The Guardian}.

\bibitem[{Yuan et~al.(2021)Yuan, Neubig, and Liu}]{yuan2021bartscore}
Weizhe Yuan, Graham Neubig, and Pengfei Liu. 2021.
\newblock Bartscore: Evaluating generated text as text generation.
\newblock \emph{Advances in neural information processing systems}, 34:27263--27277.

\end{thebibliography}

\clearpage
\appendix

\section{Information Extraction Prompt}
\label{sec:ie-prompt}
\begin{lstlisting}
<role_and_objective>
You are a legal entity information extractor. Your task is to systematically identify and extract specified entity types from provided legal text.
</role_and_objective>

<checklist>
Begin with a concise checklist (3-7 bullets) of what you will do; keep items conceptual, not implementation-level.
</checklist>

<instructions>
- Identify and extract all instances of the following entity types:
  - Date
  - Geopolitical Entity (GPE)
  - Organization (including those ending in division, department, agency, etc.)
  - Public/Private Hearing
  - In-Chamber/Virtual Hearing
  - Judge Name
  - Date of Decision
- For each detected entity instance, provide:
  - Normalized value (e.g., date in ISO format, standardized organization name, etc.), or None if normalization is not possible
  - Raw string exactly as found in the text
  - Context: A substring from the source that provides proof for the entity 
- If multiple instances of an entity are detected, list each instance as a separate object within the corresponding entity type array.
- If no instances are found for an entity type, assign None to that key.
- For incomplete or ambiguous information, set normalized to None, but still provide the raw string and context for traceability.
</instructions>

<output_format>
Return results as a JSON object structured as a list of entities.
Ensure the output strictly matches the above JSON schema, with null for any entity type not found.
</output_format>

<reasoning_steps>
Think step by step internally when extracting and normalizing entities. Do not reveal stepwise reasoning in output unless specifically asked.
</reasoning_steps>

<validation>
After entity extraction, validate that each required entity type has either a populated array or a null value, and that fields conform to the Output Format. If validation fails, self-correct and output the revised JSON.
</validation>

<verbosity>
- Output should be concise, with no extraneous explanations.
- Context field should be minimal but sufficient (entity plus one or two neighboring words).
</verbosity>

<stop_conditions>
- Return the structured JSON as soon as all entity extraction is completed and validated.
- If any entity instance is not found in the text, ensure the corresponding value is set to null.
</stop_conditions>

<notes>
- Do not include any information or commentary outside the required JSON output.
- When in doubt regarding normalization, prioritize transparency by including the raw string and context but flagging normalization as null where appropriate.
</notes>
\end{lstlisting}

\section{LLM Scorer Prompt}
\label{sec:llm-scorer-prompt}
\begin{lstlisting}
<task>
Name: Hallucination Detection
Description: Given an extracted context and an entity, determine if the entity is of the correct type and if the context supports the entity.
</task>

<examples>
    <example>
        <extracted_entity>Date: 2025-06-14</extracted_entity>
        <context>he said they would show up on jun 14, 25</context>
        <output>This would not be a hallucination. (return false)</output>
    </example>
    <example>
        <extracted_entity>Judge: Joe Burrow</extracted_entity>
        <context>the claimant joe burrow</context>
        <output>This would be a hallucination, as the context informs us that the extracted entity is a claimant and not a judge. (return true)</output>
    </example>
    <example>
        <extracted_entity>Hearing Type: In Person</extracted>
        <context></context>
        <output>This would be a hallucination as there is no context provided. (return true)</output>
    </example>
</examples>
\end{lstlisting}

\end{document}